\documentclass[letterpaper, 10 pt, conference]{ieeeconf}  

\usepackage{stfloats}
\usepackage{graphicx}

\usepackage{cite}

\usepackage[hidelinks]{hyperref}

\IEEEoverridecommandlockouts                              

\title{\LARGE \bf
Tendon-Driven Reciprocating and Non-Reciprocating Motion via Snapping Metabeams
}

\author{
Mohsen Jafarpour$^{1}$,
Ayberk Yüksek$^{1}$,
Shahab Eshghi$^{2}$,
Stanislav Gorb$^{2}$,
Edoardo Milana$^{1}$%
\thanks{$^{1}$Department of Microsystems Engineering (IMTEK) and the Cluster of Excellence livMatS @ FIT – Freiburg Center for Interactive Materials and Bioinspired Technologies, University of Freiburg, 79110 Freiburg, Germany}
\thanks{$^{2}$Department of Functional Morphology and Biomechanics, Zoological Institute, Kiel University, 24118 Kiel, Germany}
\thanks{Corresponding authors: m.jafarpour@imtek.de, milana@imtek.de}
}

\begin{document}

\maketitle
\thispagestyle{empty}
\pagestyle{empty}

\begin{abstract}

Snapping beams enable rapid geometric transitions through nonlinear instability, offering an efficient means of generating motion in soft robotic systems. In this study, a tendon-driven mechanism consisting of spiral-based metabeams was developed to exploit this principle for producing both reciprocating and non-reciprocating motion. The snapping structures were fabricated using fused deposition modeling with polylactic acid (PLA) and experimentally tested under different boundary conditions to analyze their nonlinear behavior. The results show that the mechanical characteristics, including critical forces and stability, can be tuned solely by adjusting the boundary constraints. The spiral geometry allows large reversible deformation even when made from a relatively stiff material such as PLA, providing a straightforward design concept for controllable snapping behavior. The developed mechanism was further integrated into a swimming robot, where tendon-driven fins exhibited two distinct actuation modes: reciprocating and non-reciprocating motion. The latter enabled efficient propulsion, producing a forward displacement of about 32 mm per 0.4 s cycle ($\approx$ 81 mm/s, equivalent to 0.4 body lengths per second). This study highlights the potential of geometry-driven snapping structures for efficient and programmable actuation in soft robotic systems.

\end{abstract}

\section{INTRODUCTION}

Snapping beams are structural elements that exhibit sudden changes in configuration once a critical load is reached. This phenomenon is a classic example of nonlinear instability, enabling rapid transitions between stable and transitional configurations, often accompanied by large and reversible geometric changes \cite{virgin2023elastic, hussein2020analytical, rafsanjani2016snapping}. Unlike smooth elastic deformations, snapping occurs over short timescales, enabling structures to release or store mechanical energy efficiently. The sharp nonlinear response of snapping beams can be precisely tailored through their geometry and boundary conditions, offering a simple yet efficient strategy for programming complex mechanical responses \cite{rafsanjani2016snapping, shan2015multistable, hussein2020analytical, zhang2021rotational}.

Unit cells consisting of snapping beams have been widely used in the development of mechanical metamaterials, exploited as recoverable and reusable energy-absorption mechanisms \cite{shan2015multistable, yang2019multi}, and in soft robots to achieve fast actuation, amplified motion, stiffness tunability, and adaptive deformation without relying on complex control systems \cite{zhang2022bioinspired, qing2024spontaneous, milana2025physical}.

While numerous studies have explored designs based on snapping beams with various geometries and material compositions for different purposes \cite{rafsanjani2016snapping, shan2015multistable, zhang2021rotational, yang2019multi, abbasi2023snap, zhang2022bioinspired, milana2025physical, qing2024spontaneous, yan2025snap, wu2025remorphable}, the use of soft polymers, such as thermoplastic polyurethane (TPU), for fused deposition modeling (FDM) 3D printing has become common practice, as TPU provides high elasticity, strength, and durability \cite{wallin20183d, desai2023thermoplastic}. Despite all these advantages, using TPU in dynamic applications can be challenging due to its significant rate-dependent behavior and viscosity. Loading–unloading curves of TPU typically show pronounced hysteresis and energy dissipation \cite{qi2005stress}. Although these properties are beneficial when designing metamaterials for energy absorption, they become disadvantageous in soft robotic systems, where rapid shape recovery and efficient energy release are desired. The slow recovery and energy loss associated with TPU-based snapping structures are, therefore, contrary to the very reason snapping mechanisms are employed.

The limitations imposed by the intrinsic properties of materials can sometimes be addressed through rational structural design, as achieving specific mechanical responses through geometry is the core concept behind mechanical metamaterials \cite{zadpoor2016mechanical}. Our aim is to apply this principle to develop a snapping metabeam that combines sufficient elasticity and resilience to undergo large deformations with high energy recovery. Mathematically simple, readily tunable spiral curves, previously explored as an effective geometrical strategy to increase the deformability of stiff and brittle materials such as polylactic acid (PLA) \cite{zarrinmehr2017interlocked, zhang2023greek, jafarpour2025pla}, serve as the underlying concept for developing the metabeam in this study.

Using Archimedean double-spirals as unit cells and interconnecting them in series, we develop a spiral-based metabeam. Two metabeams arranged at an angle form the snapping structure examined in this study. The models are fabricated using an FDM 3D printer with PLA filament, and their mechanical behavior is investigated under different boundary conditions and tensile loading–unloading cycles. Moreover, the deformation path and snapping motion of the structures are tracked through trajectory analysis. In the next step, the developed spiral-based metabeam is employed as a mechanism capable of exhibiting both reciprocating and non-reciprocating motion. Based on this mechanism, a swimming robot is designed and fabricated to demonstrate the functionality and applicability of the proposed snapping structure. The robot serves as a proof of concept, demonstrating how geometry-driven snap-through behavior can be utilized for efficient and controllable motion in soft robotic systems.

\section{Methods}

\subsection{Design and Fabrication}

\begin{figure}[!b]
    \centering
    \includegraphics[width=\columnwidth]{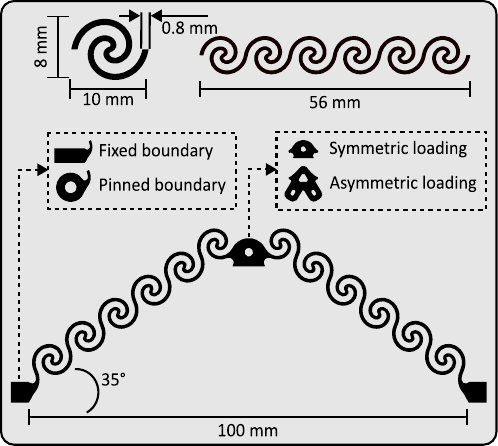}
    \caption{Design and configurations of the snapping structure. The spiral-based metabeam, consisting of interconnected double-spiral unit cells, forms the snapping structure shown under the tested boundary conditions (fixed and pinned) and loading types (symmetric and asymmetric).}
    \label{fig:Fig1}
\end{figure}

The design process started by sketching a planar double-spiral unit cell using the Double-Spiral Design software \cite{jafarpour2024double}. Using Archimedean spiral curves, an 8 mm × 10 mm unit cell was generated (Fig.~\ref{fig:Fig1}). The thickness of the coils within the double-spiral was set to 0.8 mm. Six identical double-spiral unit cells were subsequently interconnected in series to form a 56 mm-long metabeam, which was rotated by 35° and mirrored to develop the snapping structure. Finally, the planar sketch was extruded out-of-plane by 10 mm, resulting in the 3D model of the snapping structure.

The snapping structure was further modified according to the specific loading scenarios and boundary conditions considered in this study. Adjustments were made to ensure compatibility with the side constraints, including pinned and fixed boundary conditions. Additionally, to enable off-center loading, two auxiliary anchor holes were incorporated at ±4 mm from the centerline (Fig.~\ref{fig:Fig1}), allowing controlled asymmetric loading, as described in Section II-B.

The 3D models were fabricated using an FDM 3D printer (Original Prusa XL, Prusa Research, Praha, Czech Republic) with PLA filament (NX2 PLA, Extrudr \textbar{} FD3D GmbH, Lauterach, Austria) for mechanical experiments. PLA was chosen as a widely adopted 3D printing material due to its ease of processing and consistent mechanical performance \cite{dey2021review}. Its relatively high stiffness makes it suitable for demonstrating how spiral-based designs can achieve substantial deformability, induce sharp snap-through behavior, and enable rapid elastic recovery after unloading.

\subsection{Mechanical Testing}

Mechanical experiments were carried out using our ZwickiLine uniaxial testing machine (Zwick Roell, Ulm, Germany) equipped with a 10 kN load cell. All tests were performed under quasi-static conditions with a loading rate of 1~mm\,s$^{-1}$. For each model, three samples were fabricated and each tested twice to ensure statistical reliability. The experimental procedures were defined according to the purpose of the study and categorized into the following two scenarios:

\subsubsection*{1) Snapping behavior under tensile loading}
The snapping structure was tested under two boundary conditions: (i) both ends fixed (fixed–fixed) and (ii) both ends pinned (pinned–pinned). The fixed boundary condition was implemented by nesting the structure ends in a rigid fixture and fastening them with bolts to restrict motion in all directions. For the pinned boundary condition, the structure ends were connected to the fixture through pin joints, allowing in-plane rotation while preventing translational motion. In both configurations, the midpoint of the structure was connected to a rigid bar through a pin joint, with the opposite side of the rigid bar pinned to the load cell of the testing machine. The crosshead was displaced vertically upward to a maximum displacement of 55 mm, followed by unloading to the initial position (Video S1). The test setup was designed to mimic tendon-driven actuation while allowing the machine to record both the loading and unloading phases, even in the presence of bistability. The force–displacement data were used to compare the mechanical response of the structures, including their snapping force, stability, and energy trapping, dissipation, and release characteristics.

\subsubsection*{2) Trajectory analysis of the snapping behavior}
The second set of experiments was conducted to analyze the trajectory of the snapping motion when the structure was loaded off-center under three boundary conditions: (i) fixed-fixed, (ii) pinned–pinned, and (iii) fixed–pinned. The samples were mounted in the respective configurations and pulled by a rigid bar pinned to the structure once at -4 mm and once at +4 mm offset from the centerline to create a controlled asymmetry in loading. The deformation of the beams was recorded, and the motion of a small reference marker placed at the central tip of the structure was tracked throughout loading and unloading using a custom Python script based on OpenCV. The extracted coordinates were saved frame by frame and used to reconstruct the displacement trajectories and visualize the deformation process (Video S2).

For the symmetric boundary conditions (fixed–fixed and pinned–pinned), loading the holes at -4 mm and +4 mm was expected to produce deformation patterns and trajectories mirrored about the centerline. In contrast, under the asymmetric fixed–pinned boundary condition, loading the two offset holes was expected to lead to distinct behaviors. To distinguish these cases, the test in which the -4 mm hole (on the pinned side) was loaded is referred to as fixed–pinned (pin), and the test in which the +4 mm hole (on the fixed side) was loaded is referred to as fixed–pinned (fix).

\subsection{Swimming Robot}

The snapping structure made from the metabeams was used as a tendon-driven mechanism enabling both reciprocating and non-reciprocating motion. To demonstrate the functionality and applicability of the developed mechanism, a simple swimming robot was constructed (Fig.~\ref{fig:Fig4}). The robot consisted of two identical snapping structures arranged symmetrically on the sides of a 20 cm long rigid body. One end of each structure was fixed, and the other end was pinned to the main body. A 0.8 mm thick fin made from PLA was attached to the tip of each snapping structure.

Both structures, each with two holes located at ±4 mm from their centerline, were actuated simultaneously by tendons that passed through one of these holes and were driven by a small pneumatic actuator connected to an air compressor via flexible tubes, producing periodic pull–release cycles. Each actuation cycle, including tendon pulling and release, was completed in 0.4 s. The pneumatic pressure (1.5 bar) was regulated to pull the tendon until snapping occurred ($\approx$ 46 mm) and then released, resulting in effectively displacement-controlled actuation. Since both tendons were driven in phase, the two fins moved synchronously during actuation (Video S3). Two actuation modes were tested: in the fixed–pinned (fix) mode, the tendon passed through the hole on the fixed side, and in the fixed–pinned (pin) mode, the tendon passed through the hole on the pinned side.

\section{Results and Discussion}

\begin{figure}[!t]
    \centering
    \includegraphics[width=\columnwidth]{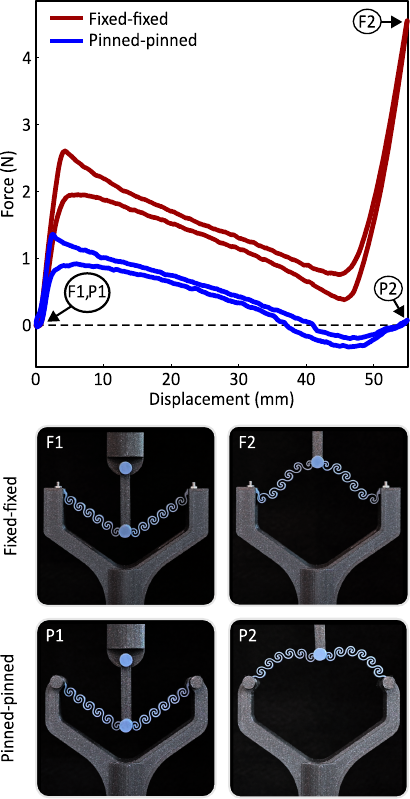}
    \caption{Representative force–displacement curves of the PLA metabeam under fixed–fixed and pinned–pinned boundary conditions, showing distinct snapping behavior. The photos illustrate the metabeams before loading and at the maximum applied displacement.}
    \label{fig:Fig2}
\end{figure}

Building on the concept of using spiral curves as a structural strategy to enhance the deformability of a stiff and brittle material such as PLA, we explored the functionality of a metabeam composed of double-spiral unit cells in the development of a snapping structure. The mechanical behavior of the developed structure was investigated experimentally to understand its nonlinear response under different boundary conditions. The force–displacement curves obtained under tensile loading reveal distinct deformation characteristics depending on the configuration. Fig.~\ref{fig:Fig2} shows representative force–displacement curves from the mechanical tests conducted on the structure under the fixed–fixed and pinned–pinned boundary conditions.

The spiral-based metabeam exhibited large, reversible deformation under both boundary conditions, confirming that spiral geometries enhance compliance by distributing strain more evenly across the structure. This provides a simple, yet effective means of achieving reversible snapping behavior, even when fabricated from relatively stiff materials.

\begin{figure*}[t!]
    \centering
    \includegraphics[width=\textwidth]{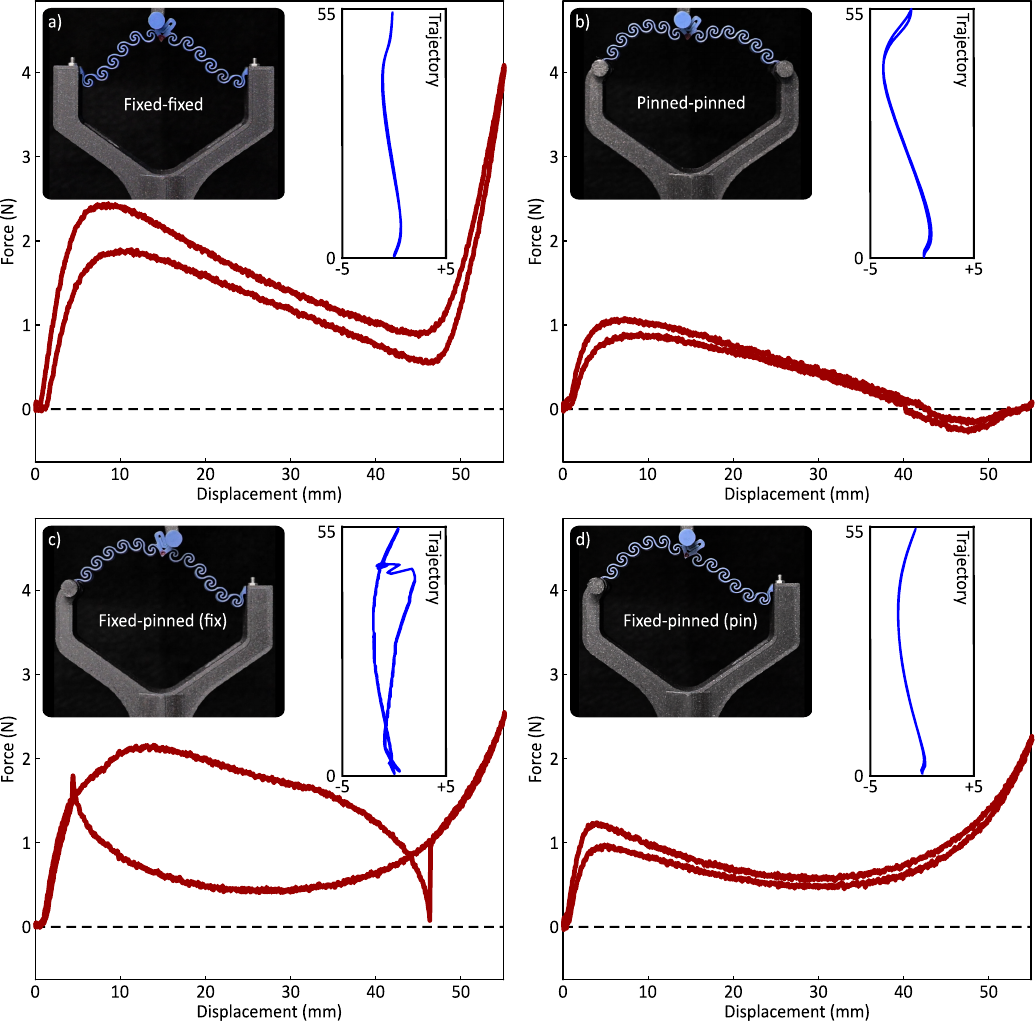}
    \caption{Mechanical behavior of the snapping structure under asymmetric tensile loading and varying boundary conditions. Representative force–displacement curves for (a) fixed–fixed, (b) pinned–pinned, and (c, d) fixed–pinned configurations with asymmetric loading applied on the fixed and pinned sides, respectively. The insets show the deformed structure at the maximum displacement (55 mm) and its tip trajectory during loading and unloading.}
    \label{fig:Fig3}
\end{figure*}

Both curves demonstrate instability-driven deformation, where the beams buckle after the critical load is reached, leading to a reduction in stiffness as the structure transitions from a stable state to a new stable or unstable configuration. Under the fixed–fixed boundary condition, the structure exhibits monostable behavior as it deforms without any change in force direction and returns to its initial configuration during unloading, with part of the input energy dissipated and the rest elastically stored during loading and subsequently released after load removal. In contrast, the structure shows bistability under the pinned–pinned boundary condition, with the force becoming negative during both loading and unloading. During loading, some of the input energy remains trapped in the structure, while the rest is dissipated or released as it settles into the second stable state \cite{shan2015multistable, yang2019multi}. During unloading, additional work must be performed to overcome the energy barrier and restore the initial configuration, at which point the previously trapped energy is released. These differences illustrate how boundary conditions influence the mechanical response of the metabeams.

Quantitatively, the fixed–fixed configuration required a higher critical force (2.74 ± 0.32 N) compared to the pinned–pinned configuration (1.24 ± 0.14 N), indicating that stronger constraints increase the resistance to snap-through. The energy dissipation ratio, defined as the ratio of dissipated energy to the work input during a loading–unloading cycle, was approximately 20\% in both cases (0.18 ± 0.03 for the fixed–fixed and 0.22 ± 0.03 for the pinned–pinned configuration), indicating similarly low hysteresis and efficient elastic energy recovery of PLA. In the fixed–fixed case, the input energy is primarily stored elastically and released during unloading, whereas in the bistable pinned–pinned structure, approximately 13\% of the total energy release occurs during loading as the structure transitions to its second stable state.

Fig.~\ref{fig:Fig3} presents the force–displacement curves obtained from the off-center (asymmetric) loading experiments under the three boundary conditions, together with the corresponding tip trajectories and snapshots of the deformed structures at their maximum displacement. For the symmetric configurations, fixed–fixed and pinned–pinned, the overall mechanical behavior remained consistent with that observed under symmetric loading. The fixed–fixed structure exhibited monostable behavior, with the force increasing to a peak value followed by a gradual decrease during deformation and a smooth recovery during unloading. The pinned–pinned configuration showed clear bistability, with the force becoming negative as the structure transitioned between two stable states. Introducing a small asymmetry in loading did not alter these general characteristics, although slight variations in force magnitude and trajectory shape were observed.

In the asymmetric fixed–pinned configuration, two load cases were examined. When the pinned side was pulled (fixed–pinned (pin)), the structure showed a monostable response, with the force increasing to a critical value followed by a gradual decrease during loading and elastic recovery during unloading. In contrast, when the fixed side was pulled (fixed–pinned (fix)), the structure exhibited a different behavior. The force first increased up to a critical point, after which the fixed beam (i.e., the beam attached to the fixed boundary) buckled, causing a reduction in force. Subsequently, as the pinned beam snapped rapidly, the force rose sharply at nearly constant displacement, indicating a snap-back phenomenon. During unloading, the same pattern was observed in reverse: the fixed beam gradually recovered toward its initial form, after which the pinned beam snapped back abruptly, releasing the rest of the stored elastic energy.

A substantial amount of energy was released during snap-back in the loading phase, resulting in a large area between the loading and unloading curves. This difference cannot be attributed to hysteresis or friction, as such effects were absent in the other configurations, which displayed only minimal energy dissipation. The large area instead reflects the release of stored elastic energy, the precise amount of which cannot be determined experimentally. Future numerical simulations will be necessary to reconstruct the detailed force–displacement relationship and quantify the energy released during the snap-back transition \cite{sun2019snap, gorissen2020inflatable}.

\begin{figure}[!b]
    \centering
    \includegraphics[width=\columnwidth]{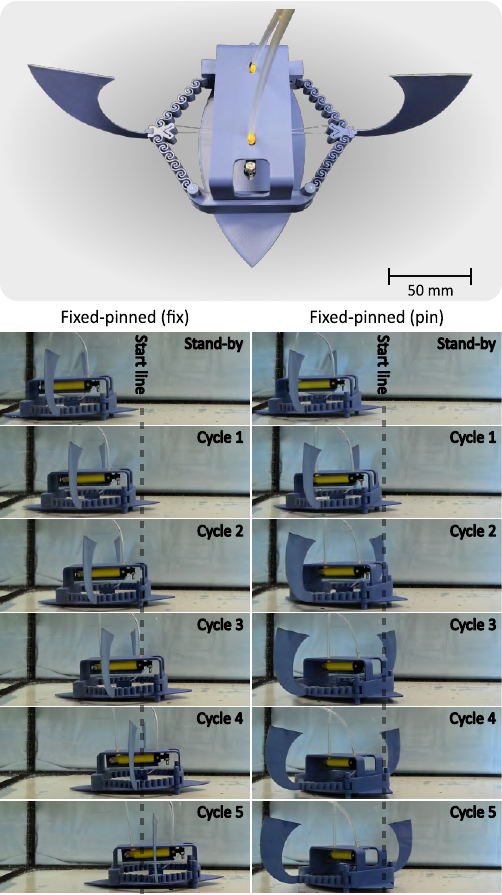}
    \caption{Swimming robot driven by tendon-actuated snapping structures. The top view shows the two identical snapping mechanisms arranged symmetrically on the sides of a rigid body. Sequential snapshots demonstrate the robot’s motion underwater during five actuation cycles with two asymmetric loading configurations: fixed–pinned (fix) and fixed–pinned (pin).}
    \label{fig:Fig4}
\end{figure}

The trajectory results complement these findings. In the fixed–fixed, pinned–pinned, and fixed–pinned (pin) configurations, the tip of the structure followed nearly identical paths during loading and unloading, demonstrating reciprocating motion and complete shape recovery (Video S2). In contrast, the fixed–pinned (fix) configuration displayed non-reciprocating motion: the loading and unloading trajectories diverged considerably, reflecting the asymmetry of the constraints and the snap-back behavior. It is noteworthy that the unloading trajectory in the fixed–pinned (fix) case closely resembled that of the fixed–pinned (pin) configuration; the difference originated primarily from the loading path, where the structure in the fixed–pinned (fix) configuration underwent asymmetric buckling and snap-back, ultimately resulting in non-reciprocating motion. This geometry-induced asymmetry represents a directional deformation mechanism later exploited in the swimming robot, whose performance is compared with that of the fixed–pinned (pin) mode.

Fig.~\ref{fig:Fig4} shows the swimming robot and the time-lapse snapshots over successive actuation cycles for the two modes. For each mode, positions at the end of cycles 1 through 5 are shown, making the net progression per cycle comparable.

During each actuation cycle, the tendons are first pulled to deform the fins and then released. The forward swimming motion occurs during the release phase, when the fins return to their undeformed state and generate a thrust (Video S3). In the fixed–pinned (fix) mode, the fins underwent non-reciprocating deformation consistent with the snap-back response, producing a directional flow and steady forward motion. The robot advanced on average $32.48 \pm 3.76~\mathrm{mm}$ per $0.4~\mathrm{s}$ cycle ($\approx 81~\mathrm{mm/s}$, equivalent to 0.4 body lengths per second), with minimal backward slip within each cycle. In contrast, in the fixed–pinned (pin) mode, the fins exhibited reciprocating deformation, and the motion within each cycle included a noticeable backward segment during tendon pulling, followed by a forward movement during release. As a result, the robot still progressed forward overall, with an average displacement per cycle of $9.61 \pm 1.30~\mathrm{mm}$, which was substantially smaller than that of the fixed--pinned (fix) mode and resulted in lower propulsion efficiency.

As the back of the robot was in contact with the wall of the water tank at the beginning of the experiments (Fig.~\ref{fig:Fig4}), the backward motion in the fixed–pinned (pin) case was mechanically restricted during the first cycle, and the robot moved forward by almost the same distance in both modes. Once the robot was released from the wall, the difference between the two modes became evident.

This simple swimming robot was developed as a proof of concept to demonstrate how geometry-driven snapping behavior can be translated into functional motion. While the design was not optimized for propulsion efficiency, it successfully highlighted the direct link between deformation patterns and locomotion performance. The results show that the employed configuration can generate consistent and directional motion solely through geometric programming.

Future studies will focus on advancing this concept beyond the current demonstrator by taking advantage of numerical simulations. A deeper understanding of the stress distribution within the double-spirals is needed to identify optimal geometries for enhanced performance and long-term durability. Numerical modeling will make it possible to analyze the unconventional snapping behaviors observed experimentally and to reconstruct complete loading–unloading curves, enabling a more accurate estimation of the energy released or dissipated during deformation. Such analyses would allow for systematic exploration of the design parameters governing the snapping response and stability, enabling further tuning of deformation modes and energy efficiency. A comparative study between spiral-based and straight-beam geometries, as well as between materials of different stiffness, such as PLA and TPU, will provide insight into how geometry and material properties jointly influence snapping dynamics. Combining experimental investigations with computational modeling will also provide a foundation for refining the current design toward more controlled and efficient motion generation in tendon-driven soft robotic systems.

\section{CONCLUSIONS}

This study introduces a tendon-driven snapping mechanism consisting of spiral-based metabeams capable of generating both reciprocating and non-reciprocating motion through geometry-driven snapping transitions. The approach combines simple monolithic fabrication with tunable mechanical response, enabling controllable motion without complex actuation or control systems. The findings highlight the potential of snapping metabeams as efficient and compact mechanisms in soft robotic applications, where motion is achieved through programmed geometric instability.

\addtolength{\textheight}{-12cm}   




\section*{ACKNOWLEDGMENT}

This work was funded by the Volkswagen Foundation under the “Pioneering Research – Exploring the Unknown Unknown” program - 9D761. Additional funding was provided by the Deutsche Forschungsgemeinschaft (DFG, German Research Foundation) under Germany’s Excellence Strategy – EXC-2193/1 – 390951807 and by the DFG – GO 995/34–2 as part of the priority program 2100 Soft Material Robotic Systems.


\bibliographystyle{IEEEtran}
\bibliography{myrefs}

\end{document}